\documentclass{article}

    \PassOptionsToPackage{numbers, compress}{natbib}

\usepackage[preprint]{neurips_2019}

\usepackage[utf8]{inputenc} 
\usepackage[T1]{fontenc}    
\usepackage{hyperref}       
\usepackage{url}            
\usepackage{booktabs}       
\usepackage{amsfonts}       
\usepackage{nicefrac}       
\usepackage{microtype}      
\usepackage{xcolor}
\usepackage{enumerate}
\usepackage{amsmath}
\usepackage{pifont}
\usepackage{floatrow}
\usepackage{graphicx}
\usepackage{float}
\usepackage{wrapfig}
\usepackage{adjustbox}
\newcommand{\error}[1]{\scalebox{0.8}{$\pm{#1}$}}

\newcommand{\cmark}{\ding{51}}%
\newcommand{\xmark}{\ding{55}}%
\newcommand*\samethanks[1][\value{footnote}]{\footnotemark[#1]}
\title{Few-shot Classification via Adaptive Attention }

\author{Zihang Jiang\thanks{Equal contribution.}~~,~
Bingyi Kang\samethanks~~, Kuangqi Zhou \& Jiashi Feng  \\
	National University of Singapore\\
	\texttt{\{jzihang, kang, kzhou\}@u.nus.edu,}
	\texttt{elefjia@nus.edu.sg} \\
}

\begin{document}

\maketitle
\begin{abstract}
Training a neural network model that can quickly adapt to a new task is highly desirable yet challenging for  few-shot learning problems. 
Recent few-shot learning methods mostly concentrate on developing various meta-learning strategies from two aspects, namely optimizing an initial model or learning a distance metric. 
In this work, we propose a novel few-shot learning method via optimizing and fast adapting  the query sample representation based on very few reference samples. To be specific, we devise a simple and efficient meta-reweighting strategy to adapt the sample representations and generate soft attention to refine the representation such that the relevant features from the query and support samples can be extracted for better few-shot classification.  Such an adaptive attention model is also able to explain what the classification model is looking for as the evidence for classification to some extent. As demonstrated experimentally, the proposed model achieves state-of-the-art classification results on various benchmark few-shot classification and fine-grained recognition datasets.\footnote{Code will be released at \url{https://github.com/zihangJiang/Adaptive-Attention}}

\end{abstract}

\section{Introduction}

In recent years,  few-shot learning problems  \cite{gidaris2018dynamic, li2019revisiting, snell2017prototypical, sung2018learning,vinyals2016matching} have attracted intensive attention, in which a model must be adapted to tackling unseen classes with only a few training examples. Humans are able to learn new concepts easily with only a few samples, but this could be highly difficult for a neural network model since it typically needs a lot of training data to extract meaningful features or otherwise it would suffer over-fitting.

To ensure a good generalization ability to novel classes, most few-shot learning methods train the model on a collection of tasks, each of which instantiates a few-shot learning problem and contains both query samples and a few support samples from different classes. Following the episodic learning paradigm~\cite{thrun2012learning}, previous works train the model by optimizing the initial one to fast adapt to various tasks, \emph{e.g.} MAML~\cite{finn2017model}, or learning a good distance metric that can cluster samples of the same class together~\cite{snell2017prototypical, vinyals2016matching, sung2018learning}.
However, MAML-alike optimization based methods usually suffer high optimization complexity.
On the other hand, though metric-based methods are simpler and generally perform better, they lack flexibility and cannot offer strong adaptation to new tasks. This is mainly because, the metric learning procedure does not exploit the relations between support and query samples directly,  and the resulted model embeds the query images into the latent space without concerning the support samples.

 When conducting few-shot classification, humans usually take several glimpses at the support and query samples, pay attention to some critical parts and make the decision~\cite{dicarlo2012does,logothetis1996visual}. 
 Such a procedure is natural and can be leveraged in meta learning. Properly modeling such a procedure of attention can help fast localize  critical parts and enhance performance of few-shot classification. 
Despite several attention mechanisms that have been shown to be effective in the fully-supervised setting \cite{wang2017residual, xiao2015application}, 
in few-shot classification, existing methods are still unable to obtain an adaptive attention map w.r.t. the few support samples accurately. 

In this work, we develop an efficient meta-reweighting strategy to adapt the representations of query samples by incorporating the representations of support samples. { Channel-level reweighting coefficients are first generated from the support features through a small neural network and then applied to query features by a channel-wise multiplication to emphasize some important feature maps}. The resulted representation could therefore contain the information from both the query and support samples.
{Based on the adapted representations after  reweighting, we further propose a new support-adaptive attention mechanism: an adaptive attention map is first generated over the query sample from the merged representation by an attention module, which  indicates the spatial location of the object within the query belonging to the class of corresponding support samples}. Then the attention map is used to refine the    representation of the query by filtering out irrelevant information w.r.t.\ the support samples before classifying the query. With such an adaptive attention mechanism, a boosted few-shot learning ability can be achieved. Meanwhile, the whole model is light-weight and easy to optimize.

To summarize, our contributions are two-fold:
\begin{enumerate}[(i)]
\item We develop a feature meta-reweighting strategy to extract and exploit the information of support examples, which is different from the traditional metric learning or optimization based meta-learning methods.

\item  We propose an attention mechanism based on the meta-reweighting strategy to localize the region of interest in query samples w.r.t.\ support samples, which helps  refine the classification performance and meanwhile explains the behavior of the model to some extent.

\end{enumerate}

\section{Related Work}
\subsection{Few-Shot Learning}

Generally, few-shot learning methods 
can be divided to meta-learning based methods and metric learning based ones. In addition, some recent works also adopt transfer leaning~\cite{Sun2018MetaTransferLF} and data augmentation~\cite{mehrotra2017generative, NIPS2018_7549} to achieve high accuracy in few-shot classification tasks, but they are based on very deep network~\cite{he2016deep, zagoruyko2016wide} which involves  more intensive pretraining and computation.

\vspace{-2mm}
\paragraph{Meta-learning based methods}
In~\cite{finn2017model}, a model-agnostic meta-learning (MAML) approach was proposed to make the model adapt to novel classes within some steps of optimization during testing with only a few samples. Instead of directly optimizing the model w.r.t. the target task, MAML-alike methods~\cite{li2017meta, rusu2018meta} tried to find a good initialization for the parameters of a model, which can then generalize to the new task by a few steps of gradient decent w.r.t. the initial parameters. Recent works~\cite{antoniou2018train} and~\cite{rusu2018meta} made some improvements on MAML to enhance model performance and stability. The work \cite{khodadadeh2018unsupervised} introduced the idea of MAML to unsupervised meta-learning and also achieved impressive results.
{Another} straightforward way for quickly acquiring new knowledge is reviewing the learned knowledge. In~\cite{graves2014neural}, a stable external memory was used for a neural network to help achieve boosted performance. 
The RNN memory-based methods like \cite{santoro2016meta} adopted an LSTM to interact with the external memory. The work \cite{ravi2016optimization} trained an LSTM-based meta-model to serve as an optimizer to another learner model. Their meta model also generated a task-common initialization of parameters for the learner model in order to quickly adapt to the test environment.

\vspace{-2mm}
\paragraph{Metric learning} 
One simple metric learning based approach is embedding images into a feature space and constraining the feature vector of the same class to be close in Euclidean distance or cosine distance. The works~\cite{vinyals2016matching, snell2017prototypical} adopted this idea and further extended it to zero-shot learning and other fields. The work~\cite{laitask} enhanced the Prototypical Network by adding a reweighting module based on the task. 
In \cite{sung2018learning}, the metric was replaced with a neural network to measure two embedded features, and decide the relations between them. This practice is much better than directly using Euclidean or cosine distance since the embedding space may be very complicated. Deep metric learning is also widely used in representation learning~\cite{hoffer2015deep}, which demonstrates good performance. Our method can also be seen as learning a deep metric that measures similarity between query and support objects. Note, if we can obtain the location of the object in the image, we are actually comparing two objects instead of two images, which is easier and more effective.

\subsection{Attention Models}
Recent work~\cite{gidaris2018dynamic} proposed an attention-based classification weight generator which achieved impressive results. Our method, however, is mainly based on the spatial attention concerning the precise location of the object in the image. Actually, in the past decades, the spatial attention mechanism was studied by the vision community with the goal of both classification and object localization. It is very challenging to learn to localize objects with only image-level labels. \cite{zhou2016learning} proposed to replace some high-level layers of a classification network by a global average pooling (GAP) \cite{lin2013network} followed by a fully connected layer to generate discriminative class activation maps (CAM). \cite{zhang2018adversarial}~improved it by using a fully convolutional network (FCN) and proved that the attention maps can be obtained by directly selecting the feature maps of the last convolution layer. However, none of these methods is applicable to few-shot setting. We extend this idea to few-shot classification in this work. Our adaptive attention module is somewhat similar to the way of getting object localization maps in~\cite{zhang2018adversarial}, but their work was proposed for localization tasks.

\section{Method}

We consider a $K$-way $n$-shot classification problem. In each few-shot classification task $\mathcal{T}$, we are given a set of support samples $\mathcal{S}=\{(x_1,y_1),\cdots,(x_N,y_N)\}$ and a batch of query samples $\mathcal{Q} = \{x_{q_1},\cdots ,x_{q_m}\}$. The samples in the support set $\mathcal{S}$ are from  $K$ categories, each of which has $n$ labeled samples, \emph{i.e.}, $N=n\times K$. The few-shot classification model is required to acquire knowledge from the support set and classify the query samples accurately.  

\subsection{Model Workflow}\label{model}
\begin{figure}[ht]
  \centering
  \setlength{\belowcaptionskip}{-0.3cm}
  \includegraphics[width=1\textwidth]{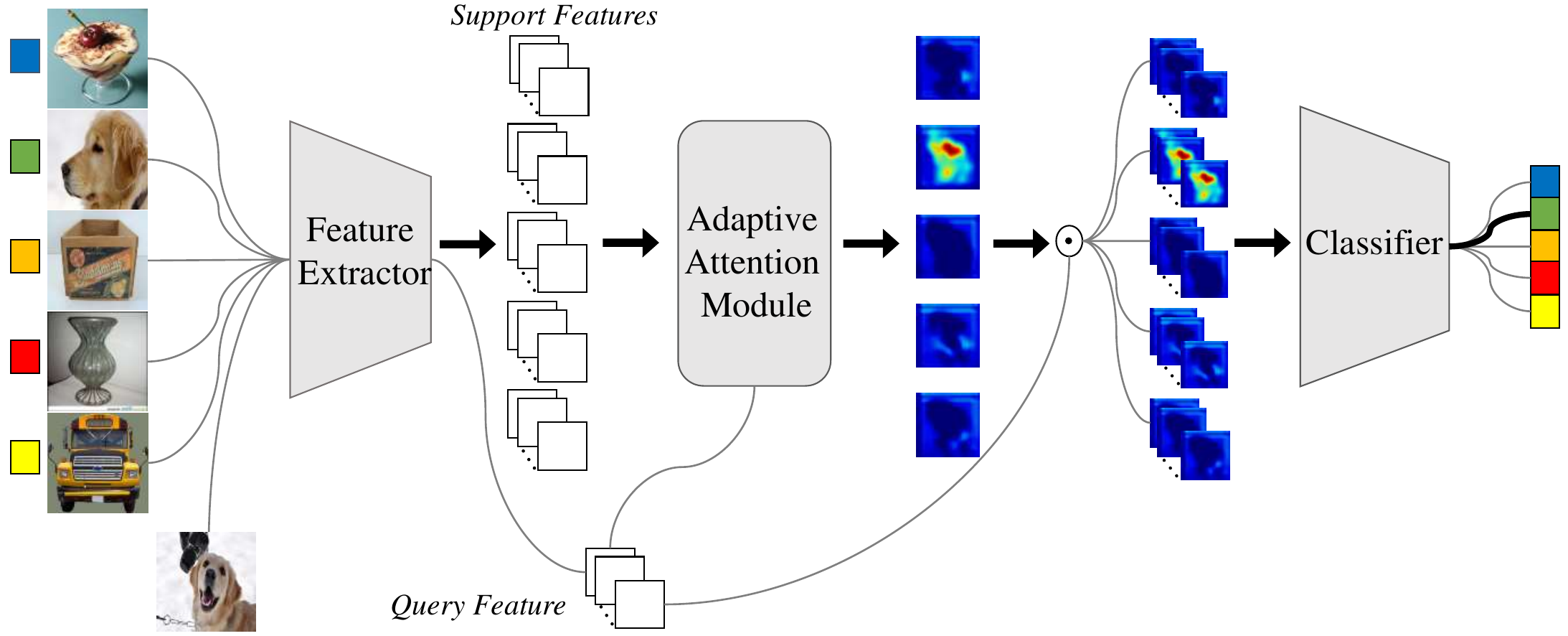}
  \caption{Illustration of our framework. The images first go
1
 through a feature extractor. Then the attention module  integrates the support  and  query feature to get adaptive attention maps. Then point-wisely multiplying   the attention maps with the query feature gives  refined query features for the classifier which outputs     the similarity scores  between the support  and the query.}
  \label{fig:model_view}
\end{figure}

An overview of our proposed  adaptive attention network 
is illustrated in Fig.~\ref{fig:model_view}. It consists of three components: a feature extractor $F(\cdot;\theta_F)$   similar to most classical few-shot models \cite{garcia2017few, gidaris2018dynamic, li2019revisiting, sung2018learning, snell2017prototypical, vinyals2016matching}, a classifier $C(\cdot;\theta_C)$   and an adaptive attention module $A(\cdot;\theta_A)$.
The classifier and  attention modules are both light-weight  as explained later.

We first explain the testing phase of the proposed model by taking  the $K$-way  1-shot classification tasks as an example. 
Both the query and support samples will go through the feature extractor $F$ at first to produce query and support  feature maps respectively. Then the adaptive attention module $A$ generates $K$ attention maps for the query feature conditioned on the $K$ support features.  Specifically, for a query feature and one of the $K$ support features, the attention module generates an attention map roughly localizing the object belonging to   the support  class in the query image. This is called an adaptive attention map for the query conditioned on one specific support. If the query image does not contain any object belonging to the class of the support image, the attention map will only highlight some background. These $K$ attention maps are then applied on the query feature to generate mask-pooled query features. These attended features are fed into the classifier $C$ to give a score indicating the confidence of the query belonging to the class of a specific support and the query is classified based on the   largest confidence  scores.
With such an adaptive attention mechanism, our model can provide better sample representation to ease the downstream classification procedure. We now proceed to explain  details of the attention module. 

\subsection{Attention Module}
\begin{figure}[t]
\setlength{\belowcaptionskip}{-0.5cm}
  \centering
  \includegraphics[width=0.9\textwidth]{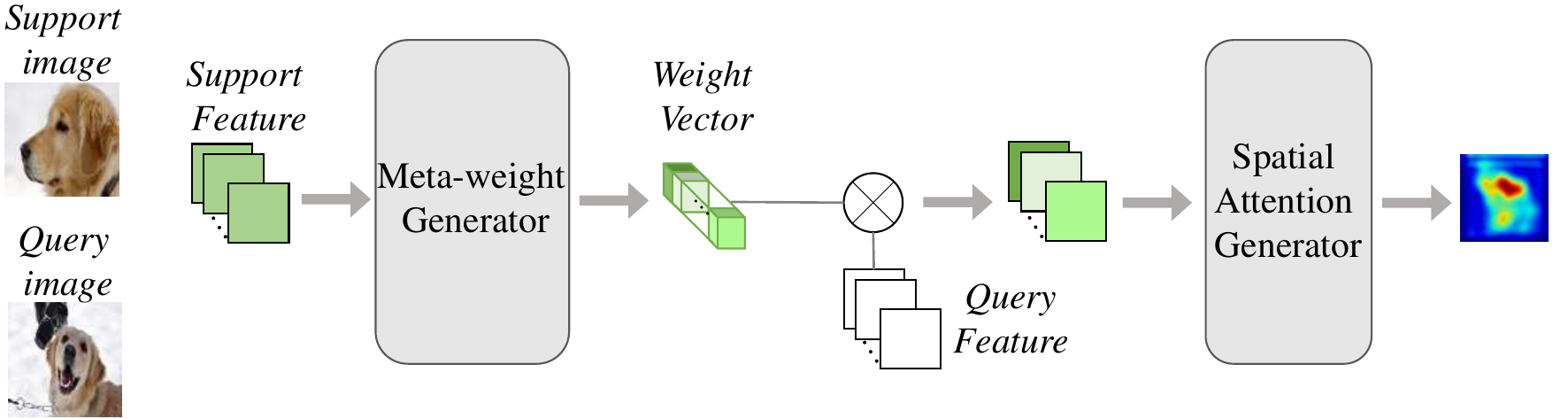}
  \caption{Illustration of our attention module. It consists of a meta-weight generator to generate class-specific weights and a spatial attention generator to precisely locate the object in query data points w.r.t. support data points.}
  \label{fig:attention_module}
\end{figure}

Taking as input a query feature and a support feature, the adaptive attention module aims to directly generate spatial  attention indicating the location of the object belonging to the support class in the query image. The key is how to adaptively generate the attention map over the query based on the information from the specific support. In this work, we propose a meta-reweighting based attention generation approach. The support features from the feature extractor are first applied to modulate the query features through channel-wise multiplication. Then the modulated query features with support information are used to generate the attention maps. In this way, the support information can be efficiently integrated into the attention generation process and the generated attention map is able to help select important regions from the query features conditioned on the support. Such an approach is beneficial to few-shot learning scenarios with only scarce support samples for training an attention model through fully supervised learning. 

As shown in Fig.~\ref{fig:attention_module}, the attention module generates the support-adaptive attention through  a meta-weight generator $A_R(\cdot; \theta_R)$ and a spatial attention generator $A_S(\cdot; \theta_S)$. 
\vspace{-3mm}
\paragraph{Meta-weight generator}
Humans recognize an object mainly based on some discriminative feature of the support examples in their brain. For example, when trying to find a cat in a room, the focus may be something hairy or with a tail.  
Motivated by this, we adopt a meta-reweighting strategy that merges the extracted features with a class-dependent weight vector. 
In particular, for a support point $x_s$ with label $y_s$, and a query point $x_q$ denote their corresponding extracted features as $f_s = F(x_s;\theta_F)$ and $f_q = F(x_q;\theta_F)$. Then we apply the meta-weight generator $A_R$ on $f_s$ to get a weight vector $w_s = A_R(f_s;\theta_R)$ for class $y_s$. The channel number of feature maps in $f_s$ equals the number of weights in $w_s$. Then we can obtain a class-specific feature 
\begin{equation}
\label{eq:reweighting}
f_q^{y_s} = f_q \otimes w_s,
\end{equation}
where $\otimes$ denotes the channel-wise multiplication. 
Note that \cite{sung2018learning} also merged the feature maps of the query and support points to get relation scores by straightly concatenating the two feature maps. Differently, our channel-wise multiplication preserves the spatial information of the query feature while emphasizing some feature maps that are crucial for classifying the support class. If $x_q$ does not belong to the class $y_s$, the emphasized feature maps do not contain useful information for class  $y_s$, in which case the classifier predicts a lower score.
\vspace{-3mm}
\paragraph{Spatial attention generator}
Our spatial attention generator $A_S(\cdot;\theta_S)$ consists of an FCN of 2 convolutional layers with one-channel final output map. 
This architecture is inspired by~\cite{zhang2018adversarial} which proposed an efficient way to get the region of interest for the deep convolutional network in weakly supervised setting.{ The main goal of this generator is to get important spatial regions in the query image w.r.t. the support sample. Taking as input the class-specific feature $f_q^{y_s}$ of a query, the output map $M_q^{y_s} = A_S(f_q^{y_s};\theta_S)$ then serves as an attention map for the corresponding support on the input query feature map and provides attention at the spatial level that helps refine it.} 

We experimentaly justify the advantage of  using such a reweighting strategy instead of directly concatenating the feature maps like Relation Network~\cite{sung2018learning} in Section~\ref{reason}. Through visualization, one will see   the output of FCN in the spatial attention generator indeed serves as an accurate attention map.
 Given a support feature $f_s$ and a query feature $f_q$, we can get a precise attention map 
\begin{equation}\label{equation:att_map}
    M_q^{y_s} = A(f_s, f_q), 
\end{equation}
which indicates the location of the object that belongs to the corresponding support class in the query.

\subsection{Classifier}
With the adaptive attention module, we can obtain the attention map for each query sample. We then 
point-wisely multiply the attention map with the query feature map. In particular, given an extracted query feature $f_q$, we can get a class-dependent attention map $M_q^{y_s}$ w.r.t. the support feature $f_s$. The feature fed into the final classifier can then be refined by the attention map to
\begin{equation}
f_q^{M_s} = f_q \odot M_q^{y_s},
\end{equation}
where $\odot$ denotes the point-wise multiplication. In this refined feature map, the region of interest is highlighted by the class-dependent attention map provided by the attention module. The final classifier will concentrate more on this region, thus get better performance.

More specifically, our classifier consists of a pooling layer followed by several linear layers, taking as input the class-dependent attention maps $M_q^{y_s}$ and the query feature $f_q$ and outputting a single score $s_{y_s}^q = C(f_q \odot M_q^{y_s})$ for each pair of query and attention map. The detailed architecture will be given in the following section. The final output is a single score representing the confidence of the query point belonging to the corresponding support class. We rely on the largest score for the final decision.

Take as input a support and a query feature denoted as $f_s$ and $f_q$. The final output score will then be $C(A(f_s, f_q)\odot f_q)$, where $A(f_s,f_q) = A_S(A_R(f_s)\otimes f_q)$ is the adaptive attention map.
Though we can use activation function like $relu$ to force the output to be positive, the function is still asymmetric. So we add its symmetric form to get
\begin{equation}\label{equation:score}
    d(f_s, f_q) = C(A(f_s, f_q)\odot f_q)+C(A(f_q, f_s)\odot f_s),
\end{equation}
as the final output score. 
The whole framework can then be reinterpreted as deep semimetric learning. 
Note that the attention map in the original form is trying to find the location of the support object in the query image, and this symmetric form can also be interpreted as finding the location of the query object in the support image, which helps improve performance when the support image contains not only objects from the support class but also other distractor classes. 

\subsection{Training}\label{strategy}
Following the episodic training scheme \cite{vinyals2016matching}, we randomly select $K$ classes from the training dataset with $n$ samples each to form the support set $N = n\times K$ examples for training. A fraction of the rest data in those $K$ classes serve as the query set. The model is then trained on these classification tasks. 

We first explain the loss function for training the attention module to produce adaptive attention maps.

Given the support set $\{x_{1,1},\cdots,x_{1,n}\}, \cdots, \{x_{K,1},\cdots,x_{K,n}\}$ with label $y_1,\cdots,y_K$ for each class and $x_q$ be a query sample with label $y_q$. As described previously, the meta-weight generator outputs a weight vector $w_{i,j}$ for each support sample $x_{i,j}$. 
By averaging the weights of each class, we can get $K$ class-specific weights $\{w_{1},\cdots, w_{k}\}$. By Eqn.~\eqref{eq:reweighting}, our model  generates $K$ class-specific features $\{f_q^{y_1},\cdots f_q^{y_K}\}$ using the class-specific weights. By feeding them to the spatial attention generator, we can get $K$ attention maps $M_q^{y_i},\ i=1,\cdots ,K$. Then by applying a global average pooling layer, we can get $K$ scores $a_i^q = \overline{M_q^{y_i}}$ indicating the confidence that query $x_q$ belongs to  class $i$. Here $\overline{M}$ denotes the average of the attention map $M$. The cross-entropy loss for the attention module is
\begin{equation}
    L_{Att} = -\log{\frac{\exp(a_j^q)}{\sum_{i=1}^K \exp(a_i^q)}},\ j = {\arg}_{y_i = y_q}\{ i\}
\end{equation}



We then explain the loss function for training the classifier and the whole model. Using Eqn.~\eqref{equation:score} Our model outputs $K$ scores $\{s_i^q|s_i^q = \frac{1}{n} \sum_{m = 1}^n d(F(x_{i,m}),F(x_q))\} , i=1,\cdots ,K$, each of which indicates the confidence of the query samples for being classified into the same category as $x_i$. The cross entropy loss is
\begin{equation}
    L_{CE} = -\log{\frac{\exp(s_j^q)}{\sum_{i=1}^K \exp(s_i^q)}},\ j = {\arg}_{y_i = y_q}\{ i\}
\end{equation}

The final loss for end-to-end training the whole model is then $L_{total} = L_{CE} + L_{Att}$.

\raggedbottom
\section{Experiment}

\subsection{Setting}
We conduct experiments to evaluate the effectiveness of our proposed model on five datasets, which include: Omniglot~\cite{lake2011one}, \textit{mini}ImageNet~\cite{vinyals2016matching}, CUB-200~\cite{WelinderEtal2010}, Stanford Dogs~\cite{KhoslaYaoJayadevaprakashFeiFei_FGVC2011}, Stanford Cars~\cite{KrauseStarkDengFei-Fei_3DRR2013}. Details of the datasets and splits are provided in the supplementary material.
Among the above five datasets, CUB-200, Stanford Dogs and Stanford Cars are originally proposed for fine-grained recognition in the fully supervised setting, and recently applied to the more challenging few shot setting~\cite{li2019revisiting, wei2018piecewise, chen2019closerfewshot}, namely fine-grained few shot classification. Unlike \textit{mini}ImageNet, the variances in these datasets are small, and each class contains only around 100 images or less. Therefore they are more challenging than the generic datasets Omniglot and \textit{mini}ImageNet since the model is forced to learn to find more accurate evidence to make a decision.

All experiments are conducted in 5-way 1-shot or 5-way 5-shot scenario. In testing, we randomly run our model for 600 episodes on each dataset except Omniglot on which we run 1,000 episodes for a fair comparison. In each episode, we randomly batch 15 query images per class to form a query set of 75 images. The classification accuracy is then calculated by averaging the accuracies of the 600 (1,000 for Omniglot) episodes. 

We adopt the commonly used 4 layers convolutional network~(\textit{Conv-64F})~\cite{snell2017prototypical, vinyals2016matching} which has 64 channels in each layer as our feature extractor. We also evaluate our method with the \textit{ResNet-256F} backbone used in~\cite{gidaris2018dynamic, li2019revisiting,mishra2017simple}, which is deeper. In order to get a more precise attention map, we remove the last max pooling layer.
As for the attention module, we use a spatial pyramid pooling (SPP) layer followed by three linear layers with $200,200$ and $64$ channels respectively for the meta-weight generator and two convolutional layers with $64$ and $1$ channels followed by a global average pooling layer for the spatial attention generator. The amount of additional parameters for the attention module is rather small, less than $1/3$ of that for the \textit{Conv-64F} feature extractor. The classifier consists of an SPP layer followed by three linear layers with $200,200$ and $1$ channels respectively. 

\subsection{Comparison with State-of-the-arts}

\begin{figure}[ht]
\setlength{\belowcaptionskip}{-0.2cm}
  \centering
  \includegraphics[width=0.9\textwidth]{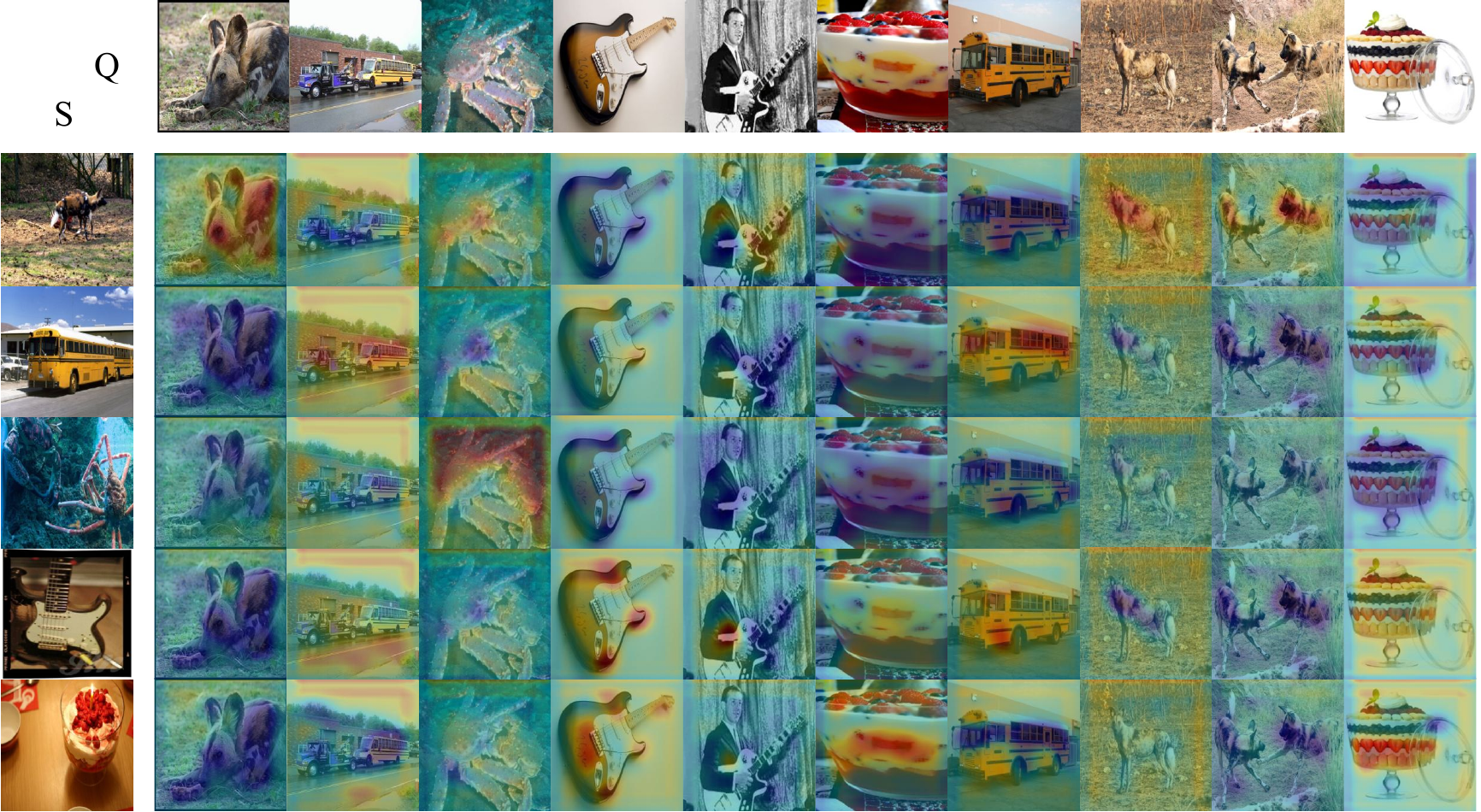}
  \caption{The adaptive attention map generated w.r.t. the support image is rendered on the query image. Top row is the query image and left column is support image. (Best viewed in color.)}
  \label{fig:attention}
\end{figure}

\paragraph{Results on generic datasets}
\begin{table}[h]
\begin{center}
\footnotesize
\begin{tabular}{l|c|c|c}
\toprule
Method &Backbone &5-way 1-shot & 5-way 5-shot\\ \midrule
Matching Network \cite{vinyals2016matching}& \textit{Conv-64F} &43.56\error{0.84} & 55.31\error{0.73}\\
ProtoNet \cite{snell2017prototypical}& \textit{Conv-64F} & 49.42\error{0.78} & 68.20\error{0.66}\\
GNN \cite{garcia2017few}& \textit{Conv-256F} &50.33\error{0.36} & 66.41\error{0.63}\\
Relation Nerwork \cite{sung2018learning} & \textit{Conv-64F} &  50.44\error{0.82} & 65.32\error{0.70} \\
DN4 \cite{li2019revisiting} & \textit{Conv-64F}& 51.24\error{0.74} & 71.02\error{0.64}\\
MAML \cite{finn2017model} & \textit{Conv-32F}&  48.70\error{1.84}& 63.11\error{0.92}\\
Dynamic-Net \cite{gidaris2018dynamic}& \textit{Conv-64F}& {56.20\error{0.86}} & {72.81\error{0.62}}\\
Dynamic-Net \cite{gidaris2018dynamic}& \textit{ResNet-256F}& 55.45\error{0.89} & 70.13\error{0.68}\\
SNAIL \cite{mishra2017simple} &\textit{ResNet-256F} &55.71\error{0.99} & 68.88\error{0.92}\\
\midrule
Ours&\textit{Conv-64F} &56.12\error{0.85}&71.48\error{0.67}\\
Ours fine-tune&\textit{Conv-64F} &56.33\error{0.85}&\textbf{72.83\error{0.67}}\\
Ours&\textit{ResNet-256F} &\textbf{59.12\error{0.91}}&{72.36\error{0.65}}\\
Ours fine-tune&\textit{ResNet-256F} &\textbf{59.26\error{0.90}}&\textbf{74.59\error{0.63}} \\
\bottomrule
\end{tabular}
\caption{Few-shot classification accuracy (\%)   with $95\%$ confidence intervals on \textit{mini}ImageNet, compared with SOTAs. For the \textit{ResNet-256F} architecture we refer to \cite{mishra2017simple}.}
\label{table:mini}
\end{center}
\end{table}
Table \ref{table:mini} reports few-shot classification performance on   \textit{mini}ImageNet.
For each task, we also use the labeled data to perform task-specific fine-tuning for one iteration and report the results referred as ``our fine-tune'' model.
We can see that our model 
outperforms all the state-of-the-art models on \textit{mini}ImageNet   for both the \textit{Conv-64F} and \textit{ResNet-256F} backbone. Noticeably, our model can offer   the   spatial location that it is looking at, as shown in Figure~\ref{fig:attention}, which largely lifts the interpretability of the model on decision making. Our model also achieves comparable performance with state-of-the-arts on Omniglot. Due to space limit, we defer the detailed results to the supplementary material.

\paragraph{Results on fine-grained datasets}
We also apply our method on fine-grained classification datasets: CUB-200, Stanford Dogs and Stanford Cars, which are more challenging. 
A classical DNN tends to suffer severe overfitting on such small datasets. We conduct similar experiments as on \textit{mini}ImageNet for both 5-way 1-shot and 5-way 5-shot scenarios. As observed from Table \ref{table:finegrain}, for 1-shot learning, our method  outperforms the state-of-the-arts  by a large margin. Such results clearly demonstrate the strong learning ability from very few shots of our proposed model and the benefits of the adaptive attention.  In the 5-shot setting, our method performs similarly to the latest method \cite{li2019revisiting}.

\begin{table}[ht]
\setlength{\belowcaptionskip}{-0.3cm}
\begin{center}
\footnotesize
\begin{tabular}{l|c|c|c|c|c|c}
\toprule
Method &\multicolumn{2}{|c|}{\textit{CUB-200}}&\multicolumn{2}{|c|}{\textit{Stanford Dogs}}&\multicolumn{2}{|c}{\textit{Stanford Cars}}\\
 &1-shot &  5-shot& 1-shot & 5-shot& 1-shot &  5-shot \\ \midrule
FSFG \cite{wei2018piecewise} & 42.10\error{1.96} & 62.48\error{1.21} & 28.78\error{2.33} & 46.92\error{2.00} & 29.63\error{2.38} & 52.28\error{1.46}\\
ProtoNet \cite{snell2017prototypical}\cite{li2019revisiting} & 37.36\error{1.00} & 45.28\error{1.03} & 37.59\error{1.00} &48.19\error{1.03}& 40.90\error{1.01} &52.93\error{1.03}\\
GNN \cite{garcia2017few}\cite{li2019revisiting} & 51.83\error{0.98} & 63.69\error{0.94}& 46.98\error{0.98} & 62.27\error{0.95} & 55.85\error{0.97} & 71.25\error{0.89} \\
DN4 \cite{li2019revisiting} & 53.15\error{0.84} & \textbf{81.90}\error{0.60}& 45.73\error{0.76}& 66.33\error{0.66} &\textbf{61.51}\error{0.85} & \textbf{89.60}\error{0.44} \\
ProtoNet \cite{snell2017prototypical}$^{*}$  & \textbf{54.54}\error{0.97} &71.02\error{0.76} & \textbf{50.57}\error{0.91} &\textbf{72.60}\error{0.68} &55.70\error{0.96} &68.68\error{0.77}\\
\midrule
Ours &\textbf{64.51}\error{0.95} & \textbf{78.62}\error{0.71} & \textbf{61.74}\error{0.98}&\textbf{77.37}\error{0.62} &\textbf{70.73}\error{0.97} & \textbf{87.72}\error{0.53}
\\ \bottomrule
\end{tabular}
\caption{Few-shot classification accuracy (\%)   with $95\%$ confidence intervals on fine-grained classification datasets, compared with SOTAs. The backbones are all \textit{Conv-64F}. Here $^*$ denotes results from our implementation.}
\label{table:finegrain}
\end{center}
\end{table}
\vspace{4mm}

\subsection{Ablation Study}
\begin{wrapfigure}[16]{r}{0.5\textwidth}
  \includegraphics[width=1\textwidth]{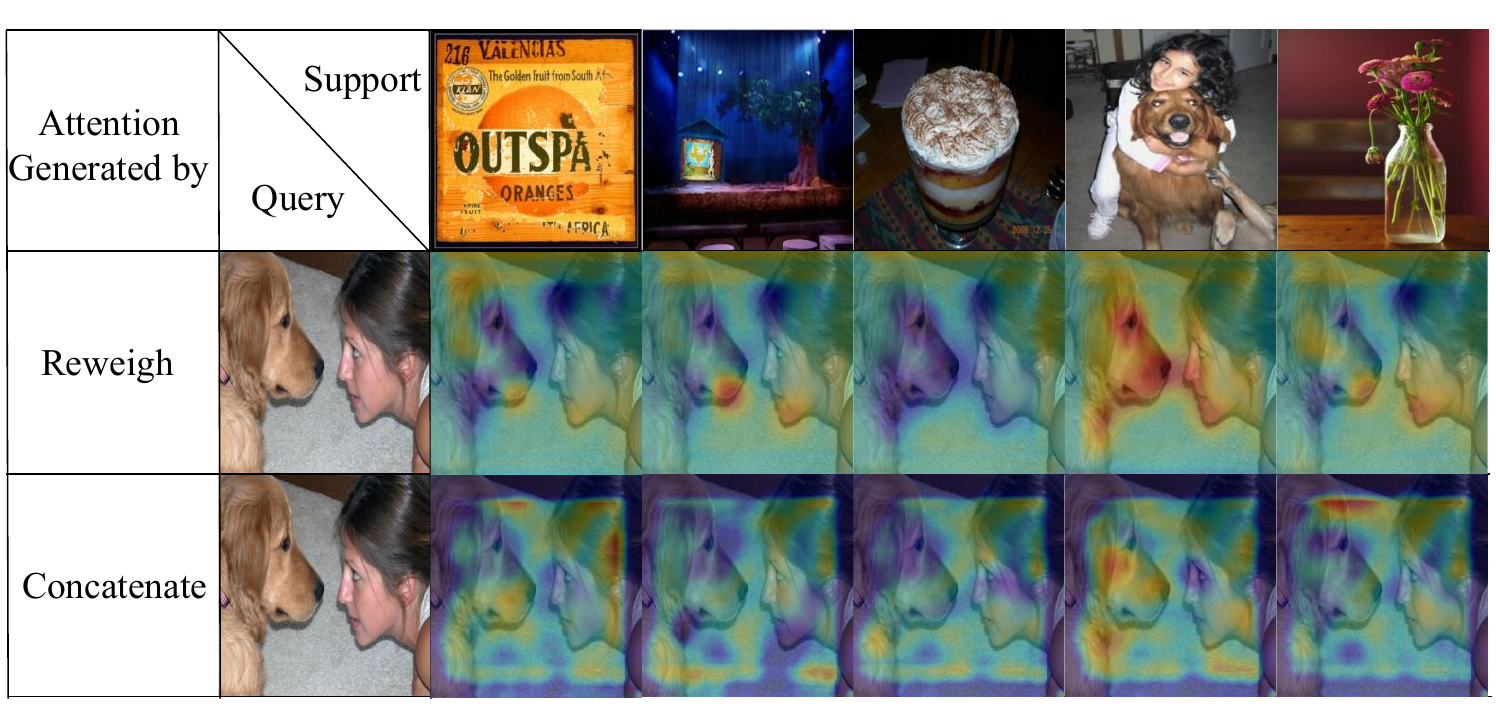}
  \caption{Comparison of the attention map generated by our meta-reweighting strategy and direct concatenation. Our attention method can  localize the dog more accurately;  the map produced by concatenation however is quite confusing. (Best viewed in color.)}
  \label{fig:compare}
\end{wrapfigure}

We perform a set of ablation studies to investigate effect of each component in our proposed model. The results are summarized in Table \ref{table:ablation}.

\vspace{-2mm}
\paragraph{Reweighting vs. concatenating for attention map generation}
\label{reason}

As shown in Table \ref{table:ablation}, replacing the meta-reweighting strategy with concatenation leads to no improvement. Note this is actually a variant of Relation Network~\cite{sung2018learning}. The attention module can act as a classifier which achieves $51.31\%$ accuracy in one-shot setting {and adding an additional classifier does not improve the model}. The generated map cannot indicate   location of the object in the query image, as shown in Figure~\ref{fig:compare}, confusing  the  classifier. The reason why directly concatenating query and support feature cannot produce a precise attention map is  straight-forward: it destroys the spatial information of the query image after concatenation. {Instead, our meta-reweighting strategy preserves the spatial information and the generated attention map precisely locates the dog as shown in the second row of Figure~\ref{fig:compare}.}

\paragraph{Effect of test data augmentation}
A key observation is that the weights generated by the meta weight generator are sometimes not so reliable since there exist distracting classes in the support images. We then apply a test data augmentation to enhance the weight generator by randomly cropping and flipping the input images and compute the mean of these vectors obtained from these augmented support data. As shown in Table \ref{table:ablation}, this stabilizes the weight generation stage and thus helps the classification. 
\begin{table}[ht]
\setlength{\belowcaptionskip}{-0.2cm}
\footnotesize
\begin{center}
\begin{tabular}{l|ccc|c|c}\toprule
 &combination & TA & AC &5-way 1-shot & 5-way 5-shot\\
\midrule
Ours&Concatenate & \xmark & \xmark & 51.31\error{0.86}&63.40\error{0.75}\\
Ours &Concatenate& \xmark & \cmark & 51.79\error{0.84}&64.13\error{0.76}\\
Ours & Reweight & \xmark & \xmark & 52.81\error{1.11}&68.91\error{0.63}\\
Ours &Reweight & \xmark & \cmark & 54.98\error{0.81} &70.02\error{0.67}\\
\midrule
Ours &Reweight& \cmark & \cmark &{56.12\error{0.85}}&{71.48\error{0.67}}\\
Ours fine-tune & Reweight & \cmark & \cmark &56.33\error{0.85}&72.83\error{0.67}\\
\bottomrule
\end{tabular}
\caption{Ablation studies on \textit{mini}ImageNet. The backbones are \textit{Conv-64F}. TA: test data augmentation; AC: the classifier component.}
\label{table:ablation}
\end{center}
\end{table}
\paragraph{Necessity of the  classifier component}
As aforementioned, the attention module can also perform the classification. To investigate its performance, we also remove the classifier and only use the attention module as a classifier. In this way, our attention module has to generate the spatial attention while performing recognition. From the results, one can observe the attention map can also be used for classification with satisfactory performance, demonstrating the adaptive attention maps indeed incorporate discriminative information from the support.  But it does not perform very well for classification compared with the additional classifier which specially focuses on classification with the help of the attention module. This demonstrates the necessity of the additional classifier.

\raggedbottom

\section{Conclusion}
In this paper, we present an efficient framework for few shot classification. It uses a meta-reweighting strategy together with an attention module to find the location of the query item w.r.t. the support samples and uses this attention map to adapatively refine query representation. 
Experimental results demonstrate the power of our proposed method especially in the one-shot setting. It outperforms all state-of-the-art models by a large margin across all real-world image datasets. Also, our method is far simpler than the recently proposed meta-learning methods which need a lot more computation and careful training. The visualization results also show potential for help  understand and improve  few-shot classification models.

\medskip

{\small
\bibliographystyle{apalike}
\bibliography{egbib}

\begin{thebibliography}{}

\bibitem[Antoniou et~al., 2018]{antoniou2018train}
Antoniou, A., Edwards, H., and Storkey, A. (2018).
\newblock How to train your maml.
\newblock {\em arXiv preprint arXiv:1810.09502}.

\bibitem[Chen et~al., 2019]{chen2019closerfewshot}
Chen, W.-Y., Liu, Y.-C., Kira, Z., Wang, Y.-C., and Huang, J.-B. (2019).
\newblock A closer look at few-shot classification.
\newblock In {\em International Conference on Learning Representations}.

\bibitem[Deng et~al., 2009]{imagenet_cvpr09}
Deng, J., Dong, W., Socher, R., Li, L.-J., Li, K., and Fei-Fei, L. (2009).
\newblock {ImageNet: A Large-Scale Hierarchical Image Database}.
\newblock In {\em CVPR09}.

\bibitem[DiCarlo et~al., 2012]{dicarlo2012does}
DiCarlo, J.~J., Zoccolan, D., and Rust, N.~C. (2012).
\newblock How does the brain solve visual object recognition?
\newblock {\em Neuron}, 73(3):415--434.

\bibitem[Finn et~al., 2017]{finn2017model}
Finn, C., Abbeel, P., and Levine, S. (2017).
\newblock Model-agnostic meta-learning for fast adaptation of deep networks.
\newblock In {\em Proceedings of the 34th International Conference on Machine
  Learning-Volume 70}, pages 1126--1135. JMLR. org.

\bibitem[Garcia and Bruna, 2017]{garcia2017few}
Garcia, V. and Bruna, J. (2017).
\newblock Few-shot learning with graph neural networks.
\newblock {\em arXiv preprint arXiv:1711.04043}.

\bibitem[Gidaris and Komodakis, 2018]{gidaris2018dynamic}
Gidaris, S. and Komodakis, N. (2018).
\newblock Dynamic few-shot visual learning without forgetting.
\newblock In {\em Proceedings of the IEEE Conference on Computer Vision and
  Pattern Recognition}, pages 4367--4375.

\bibitem[Graves et~al., 2014]{graves2014neural}
Graves, A., Wayne, G., and Danihelka, I. (2014).
\newblock Neural turing machines.
\newblock {\em arXiv preprint arXiv:1410.5401}.

\bibitem[He et~al., 2016]{he2016deep}
He, K., Zhang, X., Ren, S., and Sun, J. (2016).
\newblock Deep residual learning for image recognition.
\newblock In {\em Proceedings of the IEEE conference on computer vision and
  pattern recognition}, pages 770--778.

\bibitem[Hoffer and Ailon, 2015]{hoffer2015deep}
Hoffer, E. and Ailon, N. (2015).
\newblock Deep metric learning using triplet network.
\newblock In {\em International Workshop on Similarity-Based Pattern
  Recognition}, pages 84--92. Springer.

\bibitem[Khodadadeh et~al., 2018]{khodadadeh2018unsupervised}
Khodadadeh, S., B{\"o}l{\"o}ni, L., and Shah, M. (2018).
\newblock Unsupervised meta-learning for few-shot image and video
  classification.
\newblock {\em arXiv preprint arXiv:1811.11819}.

\bibitem[Khosla et~al., 2011]{KhoslaYaoJayadevaprakashFeiFei_FGVC2011}
Khosla, A., Jayadevaprakash, N., Yao, B., and Fei-Fei, L. (2011).
\newblock Novel dataset for fine-grained image categorization.
\newblock In {\em First Workshop on Fine-Grained Visual Categorization, IEEE
  Conference on Computer Vision and Pattern Recognition}, Colorado Springs, CO.

\bibitem[Krause et~al., 2013]{KrauseStarkDengFei-Fei_3DRR2013}
Krause, J., Stark, M., Deng, J., and Fei-Fei, L. (2013).
\newblock 3d object representations for fine-grained categorization.
\newblock In {\em 4th International IEEE Workshop on 3D Representation and
  Recognition (3dRR-13)}, Sydney, Australia.

\bibitem[Lai et~al., 2018]{laitask}
Lai, N., Kan, M., Shan, S., and Chen, X. (2018).
\newblock Task-adaptive feature reweighting for few shot classification.
\newblock {\em ACCV}.

\bibitem[Lake et~al., 2011]{lake2011one}
Lake, B., Salakhutdinov, R., Gross, J., and Tenenbaum, J. (2011).
\newblock One shot learning of simple visual concepts.
\newblock In {\em Proceedings of the Annual Meeting of the Cognitive Science
  Society}, volume~33.

\bibitem[Li et~al., 2019]{li2019revisiting}
Li, W., Wang, L., Xu, J., Huo, J., Gao, Y., and Luo, J. (2019).
\newblock Revisiting local descriptor based image-to-class measure for few-shot
  learning.
\newblock {\em arXiv preprint arXiv:1903.12290}.

\bibitem[Li et~al., 2017]{li2017meta}
Li, Z., Zhou, F., Chen, F., and Li, H. (2017).
\newblock Meta-sgd: Learning to learn quickly for few-shot learning.
\newblock {\em arXiv preprint arXiv:1707.09835}.

\bibitem[Lin et~al., 2013]{lin2013network}
Lin, M., Chen, Q., and Yan, S. (2013).
\newblock Network in network.
\newblock {\em arXiv preprint arXiv:1312.4400}.

\bibitem[Logothetis and Sheinberg, 1996]{logothetis1996visual}
Logothetis, N.~K. and Sheinberg, D.~L. (1996).
\newblock Visual object recognition.
\newblock {\em Annual review of neuroscience}, 19(1):577--621.

\bibitem[Mehrotra and Dukkipati, 2017]{mehrotra2017generative}
Mehrotra, A. and Dukkipati, A. (2017).
\newblock Generative adversarial residual pairwise networks for one shot
  learning.
\newblock {\em arXiv preprint arXiv:1703.08033}.

\bibitem[Mishra et~al., 2017]{mishra2017simple}
Mishra, N., Rohaninejad, M., Chen, X., and Abbeel, P. (2017).
\newblock A simple neural attentive meta-learner.
\newblock {\em arXiv preprint arXiv:1707.03141}.

\bibitem[Ravi and Larochelle, 2016]{ravi2016optimization}
Ravi, S. and Larochelle, H. (2016).
\newblock Optimization as a model for few-shot learning.
\newblock {\em ICLR17}.

\bibitem[Rusu et~al., 2018]{rusu2018meta}
Rusu, A.~A., Rao, D., Sygnowski, J., Vinyals, O., Pascanu, R., Osindero, S.,
  and Hadsell, R. (2018).
\newblock Meta-learning with latent embedding optimization.
\newblock {\em arXiv preprint arXiv:1807.05960}.

\bibitem[Santoro et~al., 2016]{santoro2016meta}
Santoro, A., Bartunov, S., Botvinick, M., Wierstra, D., and Lillicrap, T.
  (2016).
\newblock Meta-learning with memory-augmented neural networks.
\newblock In {\em International conference on machine learning}, pages
  1842--1850.

\bibitem[Schwartz et~al., 2018]{NIPS2018_7549}
Schwartz, E., Karlinsky, L., Shtok, J., Harary, S., Marder, M., Kumar, A.,
  Feris, R., Giryes, R., and Bronstein, A. (2018).
\newblock Delta-encoder: an effective sample synthesis method for few-shot
  object recognition.
\newblock In Bengio, S., Wallach, H., Larochelle, H., Grauman, K.,
  Cesa-Bianchi, N., and Garnett, R., editors, {\em Advances in Neural
  Information Processing Systems 31}, pages 2845--2855. Curran Associates, Inc.

\bibitem[Snell et~al., 2017]{snell2017prototypical}
Snell, J., Swersky, K., and Zemel, R. (2017).
\newblock Prototypical networks for few-shot learning.
\newblock In {\em Advances in Neural Information Processing Systems}, pages
  4077--4087.

\bibitem[Sun et~al., 2018]{Sun2018MetaTransferLF}
Sun, Q., Liu, Y., Chua, T.-S., and Schiele, B. (2018).
\newblock Meta-transfer learning for few-shot learning.
\newblock {\em CoRR}, abs/1812.02391.

\bibitem[Sung et~al., 2018]{sung2018learning}
Sung, F., Yang, Y., Zhang, L., Xiang, T., Torr, P.~H., and Hospedales, T.~M.
  (2018).
\newblock Learning to compare: Relation network for few-shot learning.
\newblock In {\em Proceedings of the IEEE Conference on Computer Vision and
  Pattern Recognition}, pages 1199--1208.

\bibitem[Thrun and Pratt, 2012]{thrun2012learning}
Thrun, S. and Pratt, L. (2012).
\newblock {\em Learning to learn}.
\newblock Springer Science \& Business Media.

\bibitem[Vinyals et~al., 2016]{vinyals2016matching}
Vinyals, O., Blundell, C., Lillicrap, T., Wierstra, D., et~al. (2016).
\newblock Matching networks for one shot learning.
\newblock In {\em Advances in neural information processing systems}, pages
  3630--3638.

\bibitem[Wang et~al., 2017]{wang2017residual}
Wang, F., Jiang, M., Qian, C., Yang, S., Li, C., Zhang, H., Wang, X., and Tang,
  X. (2017).
\newblock Residual attention network for image classification.
\newblock {\em arXiv preprint arXiv:1704.06904}.

\bibitem[Wei et~al., 2018]{wei2018piecewise}
Wei, X.-S., Wang, P., Liu, L., Shen, C., and Wu, J. (2018).
\newblock Piecewise classifier mappings: Learning fine-grained learners for
  novel categories with few examples.
\newblock {\em arXiv preprint arXiv:1805.04288}.

\bibitem[Welinder et~al., 2010]{WelinderEtal2010}
Welinder, P., Branson, S., Mita, T., Wah, C., Schroff, F., Belongie, S., and
  Perona, P. (2010).
\newblock {Caltech-UCSD Birds 200}.
\newblock Technical Report CNS-TR-2010-001, California Institute of Technology.

\bibitem[Xiao et~al., 2015]{xiao2015application}
Xiao, T., Xu, Y., Yang, K., Zhang, J., Peng, Y., and Zhang, Z. (2015).
\newblock The application of two-level attention models in deep convolutional
  neural network for fine-grained image classification.
\newblock In {\em Proceedings of the IEEE Conference on Computer Vision and
  Pattern Recognition}, pages 842--850.

\bibitem[Zagoruyko and Komodakis, 2016]{zagoruyko2016wide}
Zagoruyko, S. and Komodakis, N. (2016).
\newblock Wide residual networks.
\newblock {\em arXiv preprint arXiv:1605.07146}.

\bibitem[Zhang et~al., 2018]{zhang2018adversarial}
Zhang, X., Wei, Y., Feng, J., Yang, Y., and Huang, T.~S. (2018).
\newblock Adversarial complementary learning for weakly supervised object
  localization.
\newblock In {\em Proceedings of the IEEE Conference on Computer Vision and
  Pattern Recognition}, pages 1325--1334.

\bibitem[Zhou et~al., 2016]{zhou2016learning}
Zhou, B., Khosla, A., Lapedriza, A., Oliva, A., and Torralba, A. (2016).
\newblock Learning deep features for discriminative localization.
\newblock In {\em Proceedings of the IEEE conference on computer vision and
  pattern recognition}, pages 2921--2929.

\end{thebibliography}
}
\pagebreak
\section{Datasets}
\paragraph{Omniglot~\cite{lake2011one}}
It is a hand-written character based dataset, and we apply the split and augmentation policy in~\cite{vinyals2016matching}. The original 1,623 classes are augmented to new classes through $90^{\circ}$, $180^{\circ}$ and $270^{\circ}$
rotations. The 1,200 original classes and those after rotations are used for training, and the remaining 423 classes plus their rotated images are used for testing. All the images are resized to $28\times 28$. 
\vspace{-3mm}
\paragraph{\textit{mini}ImageNet~\cite{vinyals2016matching}}
It is a mini version of ImageNet \cite{imagenet_cvpr09} as a benchmark for few shot classification. It contains 60,000 color images from 100 classes, with 600 images in each class. Following the splits used in \cite{ravi2016optimization}, we use 64 classes for training, 16 classes for validation and the remaining 20 classes for testing. All images are resized to $84\times 84$.
\vspace{-3mm}
\paragraph{CUB-200~\cite{WelinderEtal2010}}
The Caltech-UCSD Birds 200 (CUB-200) is a dataset with images of 200 bird species. It contains $6,033$ images in total. Following the split in \cite{li2019revisiting}, we use 130 classes for training, 20 classes for validation, and the remaining 50 classes for testing respectively.
\vspace{-3mm}
\paragraph{Stanford Dogs~\cite{KhoslaYaoJayadevaprakashFeiFei_FGVC2011}}
This dataset is also a subset of ImageNet. It contains 20,580 images of 120 breeds (classes) of dogs. Similarly, we use the split in~\cite{li2019revisiting} to get 70, 20 and 30 classes for training, validation and testing respectively.
\vspace{-3mm}
\paragraph{Stanford Cars~\cite{KrauseStarkDengFei-Fei_3DRR2013}}
This dataset contains 16,185 images of 196 classes of cars. Following \cite{li2019revisiting}, we take 130, 17 and 49 classes for training, validation and testing respectively.

\section{Results on Omniglot}
Following \cite{snell2017prototypical}, we train the model on the 60-way 1-shot and 60-way 5-shot setting, and provide result for 5-way 1-shot and 5-way 5-shot scenarios in Table \ref{table:omniglot}.

\begin{table}[ht]
\setlength{\belowcaptionskip}{-0.5cm}
\begin{center}
\begin{tabular}{l|c|c}\toprule
Method &5-way 1-shot & 5-way 5-shot\\
\midrule
Matching Network\cite{vinyals2016matching} & 98.1 & 98.9\\
Prototypical Network\cite{snell2017prototypical}& 98.8&99.7\\
GNN\cite{garcia2017few}&99.2&99.7\\
Relation Network\cite{sung2018learning}&99.6\error{0.2}&99.8\error{0.1}\\
MAML\cite{finn2017model} &98.4\error{0.4}&99.9\error{0.1}\\
\midrule
ours & 99.2\error{0.1} & 99.7\error{0.1}\\
\bottomrule
\end{tabular}
\caption{5-way 1-shot and 5-way 5-shot classification accuracy(\%) results with $95\%$ confidence intervals on Omniglot compare with other state or the art method, the backbone used are \textit{Conv-64F}.}
\label{table:omniglot}
\end{center}
\end{table}

\end{document}